\documentclass{article}
\usepackage{authblk}
\usepackage{amsmath, amssymb}
\usepackage{graphicx}
\usepackage{hyperref}
\usepackage{tikz}
\usetikzlibrary{arrows.meta, positioning, shapes.geometric, backgrounds}
\usepackage{booktabs}
\usepackage{geometry}
\title{SafeRL-Lite: A Lightweight, Explainable, and Constrained Reinforcement Learning Library}
\author[1]{Satyam Mishra}
\author[2]{Phung Thao Vi}
\author[3]{Shivam Mishra}
\author[4]{Vishwanath Bijalwan}
\author[5]{Vijay Bhaskar Semwal}
\author[6]{Abdul Manan Khan}
\affil[1]{\texttt{satyam.entrprnr@gmail.com} \\Vision Mentors Ltd., Hanoi, Vietnam}
\affil[2,3]{\texttt{Vietnam National University, Hanoi}}
\affil[4]{\texttt{School of Computer Science \& AI, SR University, Warangal, India}}
\affil[5]{\texttt{Department of Computer Science, MANIT, Bhopal, India}}
\affil[6]{\texttt{University of West London, London, UK}}

\date{}

\begin{document}

\maketitle

\begin{abstract}
We introduce \textbf{SafeRL-Lite}, an open-source Python library for building reinforcement learning (RL) agents that are both \emph{constrained} and \emph{explainable}. Existing RL toolkits often lack native mechanisms for enforcing hard safety constraints or producing human-interpretable rationales for decisions. SafeRL-Lite provides modular wrappers around standard Gym environments and deep Q-learning agents to enable: (i) safety-aware training via constraint enforcement, and (ii) real-time post-hoc explanation via SHAP values and saliency maps. The library is lightweight, extensible, and installable via \texttt{pip}, and includes built-in metrics for constraint violations. We demonstrate its effectiveness on constrained variants of CartPole and provide visualizations that reveal both policy logic and safety adherence. The full codebase is available at: \url{https://github.com/satyamcser/saferl-lite}.
\end{abstract}

\section{Introduction}

Reinforcement Learning (RL) has achieved remarkable success across a wide range of domains, from game playing to robotic control and autonomous decision-making. However, the deployment of RL agents in real-world safety-critical applications remains a significant challenge due to two key limitations: (1) the lack of safety guarantees during exploration and policy execution, and (2) the opaqueness of learned policies, which hinders human understanding and trust.

In practical domains such as autonomous driving, industrial automation, and clinical decision support, agents are often required to operate under hard constraints: for example, to avoid collisions, respect velocity limits, or obey medical safety protocols. Standard RL algorithms, such as Deep Q-Networks (DQN), are typically designed to maximize cumulative reward without any explicit notion of constraint satisfaction. Violations of such constraints can lead to catastrophic outcomes, rendering these agents unusable in safety-sensitive contexts.

Furthermore, most RL models are inherently black-box in nature. Despite their impressive performance, they offer little insight into the rationale behind their decisions. As a result, when these models fail, either due to unexpected conditions or spurious correlations, it becomes nearly impossible for practitioners to diagnose and correct errors. This lack of transparency poses a barrier to adoption in regulated industries, where accountability and interpretability are non-negotiable.

To address these challenges, we introduce \textbf{SafeRL-Lite}: a lightweight, modular, and extensible Python library that enables the construction of RL agents that are both \emph{constrained} and \emph{explainable}, without requiring major modifications to standard training pipelines. SafeRL-Lite wraps around Gym environments and DQN agents to inject safety-awareness and interpretability capabilities with minimal overhead. Its contributions are threefold:

\begin{itemize}
  \item \textbf{Safety Constraints via Environment Wrappers:} Users can enforce logical or numerical constraints (e.g., "velocity must not exceed $v_{\max}$") using configurable wrappers. These wrappers monitor constraint violations in real-time and optionally terminate episodes or penalize unsafe actions.
  
  \item \textbf{Built-in Explainability:} SafeRL-Lite integrates two widely used model explanation techniques: SHAP (SHapley Additive exPlanations) and saliency maps, to attribute decision importance to input features, visualize attention over time, and trace safety-violating decisions.
  
  \item \textbf{Plug-and-Play Architecture:} The library is fully compatible with standard Gym environments and DQN training loops. It can be installed via PyPI, used in Jupyter notebooks, and extended with custom environments or constraint definitions.
\end{itemize}

With a minimal learning curve and clean API, SafeRL-Lite aims to support researchers, educators, and practitioners in building RL agents that are not only reward-optimal, but also safe, interpretable, and trustworthy.

\section{Related Work}

\textbf{Safe Reinforcement Learning (Safe RL)} focuses on optimizing policies under safety constraints, commonly formalized via Constrained Markov Decision Processes (CMDPs). Notable methods include Constrained Policy Optimization (CPO) \cite{achiam2017cpo}, Lyapunov-based constraints \cite{chow2018lyapunov}, and comparative benchmarks on safe exploration \cite{ray2019benchmarking}. These approaches provide theoretical guarantees but often require custom environments or complex solvers.

\textbf{Explainable Reinforcement Learning (XRL)} seeks to make decision-making interpretable to human users. Early work such as saliency visualizations in Atari \cite{greydanus2018visualizing} inspired further research into local explanation methods for RL agents. Recent advances utilize uncertainty-aware DQNs combined with visual XAI components like heatmaps and Q-value overlays \cite{peixoto2024xai}, pushing the field toward high-stakes applications.

However, most of these solutions remain tightly coupled with research prototypes or require heavy modification of the base RL agent. \textbf{SafeRL-Lite} bridges this gap by offering constraint wrappers and SHAP/saliency-based visualization in a modular plug-and-play design, enabling both safe training and transparent inference with minimal configuration overhead.

\section{Problem Formulation}

We consider a standard Markov Decision Process (MDP) defined by the tuple $\mathcal{M} = (\mathcal{S}, \mathcal{A}, P, r, \gamma)$, where:
\begin{itemize}
  \item $\mathcal{S}$ is the state space,
  \item $\mathcal{A}$ is the action space,
  \item $P(s'|s,a)$ is the transition probability,
  \item $r(s,a)$ is the reward function,
  \item $\gamma \in [0,1)$ is the discount factor.
\end{itemize}

A policy $\pi: \mathcal{S} \rightarrow \Delta(\mathcal{A})$ maps states to probability distributions over actions. In safe reinforcement learning, we impose a set of $m$ safety constraints $C_i(s, a) \leq 0$, such as limit on acceleration, energy usage, or violation count.

The constrained optimization problem is defined as:
\[
\max_{\pi} \mathbb{E}_{\pi} \left[ \sum_{t=0}^\infty \gamma^t r(s_t, a_t) \right] \quad \text{s.t.} \quad \mathbb{E}_{\pi} \left[ \sum_{t=0}^\infty \gamma^t C_i(s_t, a_t) \right] \leq d_i \quad \forall i = 1, \dots, m
\]
where $d_i$ denotes the acceptable threshold for constraint $i$.

\subsection{Modular Safety Wrapper}
In \texttt{SafeRL-Lite}, these constraints are not encoded into the optimization but are enforced as runtime wrappers:
\[
a_t^\text{safe} = 
\begin{cases}
a_t & \text{if } C_i(s_t, a_t) \leq 0 \ \forall i \\
\arg\min_{a' \in \mathcal{A}} \sum_i \max(0, C_i(s_t, a')) & \text{otherwise}
\end{cases}
\]
This preserves compatibility with unconstrained off-the-shelf agents such as DQNs, while still enforcing runtime safety guarantees.

\subsection{Explainability Objective: SHAP Attribution}

To support post-hoc interpretability, we compute feature attributions using SHAP (SHapley Additive exPlanations). Let $f$ denote the Q-value function $f(s) = Q(s, a)$ for a fixed action $a$. We decompose $f$ over a feature set $\mathcal{F} = \{1, \dots, d\}$.

For each input feature $j \in \mathcal{F}$, the SHAP value $\phi_j$ is defined as:
\[
\phi_j = \sum_{S \subseteq \mathcal{F} \setminus \{j\}} \frac{|S|!(d - |S| - 1)!}{d!} \left[ f(S \cup \{j\}) - f(S) \right]
\]
where:
\begin{itemize}
  \item $S$ is a subset of input features,
  \item $f(S)$ denotes the Q-value prediction with missing features in $S^c$ marginalized out,
  \item $\phi_j$ reflects the average marginal contribution of feature $j$ to $f$.
\end{itemize}

In practice, exact computation is exponential in $d$, so we use approximation strategies such as KernelSHAP.

\subsection{Joint Safety + Explainability Objective}
The proposed \texttt{SafeRL-Lite} framework therefore enables:
\begin{enumerate}
  \item \textbf{Safe control:} via action masking or runtime projection onto the safe set.
  \item \textbf{Transparent decision-making:} via real-time $\phi_j$ heatmaps per timestep.
\end{enumerate}

This hybrid formulation enables us to train unconstrained agents and still deploy them in safety-critical applications with human-aligned explanations.

\section{Library Design}

\texttt{SafeRL-Lite} is designed as a modular reinforcement learning extension library for Python, enabling constraint enforcement and interpretability without requiring changes to the agent's internal learning loop. Its lightweight architecture is compatible with any Gym environment and DQN-based policy. Below we describe the major components and their integration.

\subsection{System Architecture Overview}

\begin{figure}[h]
\centering
\begin{tikzpicture}[node distance=1.7cm and 2cm, every node/.style={font=\small}]
  \node[draw, rectangle, minimum width=2.6cm, minimum height=1.2cm, fill=blue!10] (env) {Gym Environment};
  \node[draw, rectangle, right=of env, minimum width=2.8cm, minimum height=1.2cm, fill=green!10] (wrapper) {SafeEnvWrapper};
  \node[draw, rectangle, right=of wrapper, minimum width=2.6cm, minimum height=1.2cm, fill=orange!15] (agent) {Constrained DQN Agent};
  \node[draw, rectangle, below=of agent, minimum width=2.6cm, minimum height=1.2cm, fill=purple!10] (explainer) {Explainability Module};
  \node[draw, rectangle, below=of wrapper, minimum width=2.8cm, minimum height=1.2cm, fill=gray!10] (log) {Violation Logs};

  \draw[->] (env) -- (wrapper) node[midway, above] {\scriptsize $s_t, r_t$};
  \draw[->] (wrapper) -- (agent) node[midway, above] {\scriptsize filtered $s_t$};
  \draw[->] (agent) -- (wrapper) node[midway, below] {\scriptsize action $a_t$};
  \draw[->] (agent) -- (explainer) node[midway, right] {\scriptsize $s_t, a_t$};
  \draw[->] (wrapper) -- (log) node[midway, left] {\scriptsize violation?};

\end{tikzpicture}
\caption{SafeRL-Lite architecture: the SafeEnvWrapper filters unsafe actions and observations, while attribution modules explain decisions.}
\label{fig:architecture}
\end{figure}
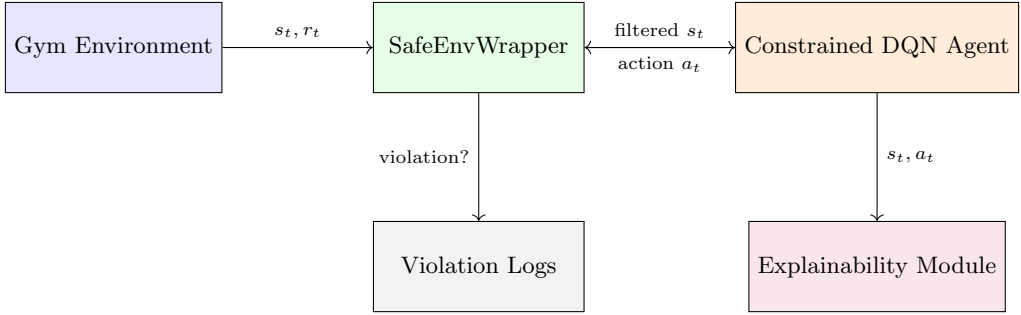

Figure~\ref{fig:architecture} depicts the core data flow between the environment, agent, and explainability modules.

\subsection{SafeEnvWrapper: Constraint Enforcement}

The \texttt{SafeEnvWrapper} wraps any Gym environment and acts as a runtime filter:
\begin{itemize}
  \item It evaluates user-defined constraints $C_i(s,a) \leq 0$ at each step.
  \item Unsafe actions are blocked, masked, or replaced.
  \item Safety violations are logged and optionally penalized in the reward signal.
  \item If configured, constraint violations trigger early episode termination.
\end{itemize}
This design does not require modifying the original environment, supporting clean layering of logic.

\subsection{ConstrainedDQNAgent: RL with Safety Awareness}

We extend a DQN agent to form the \texttt{ConstrainedDQNAgent}, which:
\begin{itemize}
  \item Accepts filtered observations from the \texttt{SafeEnvWrapper}.
  \item Logs unsafe actions and updates violation counters.
  \item Optionally applies a constraint-aware reward:
  \[
  r'_t = r_t - \lambda \sum_i \mathbb{I}[C_i(s_t, a_t) > 0]
  \]
  where $\lambda$ is a user-defined penalty weight.
\end{itemize}
This allows the agent to balance reward maximization and safety adherence during training.

\subsection{Explainability Modules: SHAP \& Saliency Attribution}

\texttt{SafeRL-Lite} supports two plug-in interpretability modules for real-time or offline explanation of agent decisions.

\begin{figure}[h]
\centering
\begin{tikzpicture}[node distance=1.7cm and 1.8cm, every node/.style={font=\small}]
  \node[draw, rectangle, fill=orange!20] (agent) {Trained Q-network};
  \node[draw, rectangle, below left=of agent, fill=purple!15] (shap) {SHAPExplainer};
  \node[draw, rectangle, below right=of agent, fill=blue!10] (saliency) {SaliencyExplainer};
  \node[draw, rectangle, below=of shap, fill=gray!10] (viz1) {Bar Plot / Heatmap};
  \node[draw, rectangle, below=of saliency, fill=gray!10] (viz2) {Gradient Map};

  \draw[->] (agent) -- (shap);
  \draw[->] (agent) -- (saliency);
  \draw[->] (shap) -- (viz1);
  \draw[->] (saliency) -- (viz2);
\end{tikzpicture}
\caption{Explainability pipeline for SHAP and Saliency-based visual attributions.}
\label{fig:explain}
\end{figure}
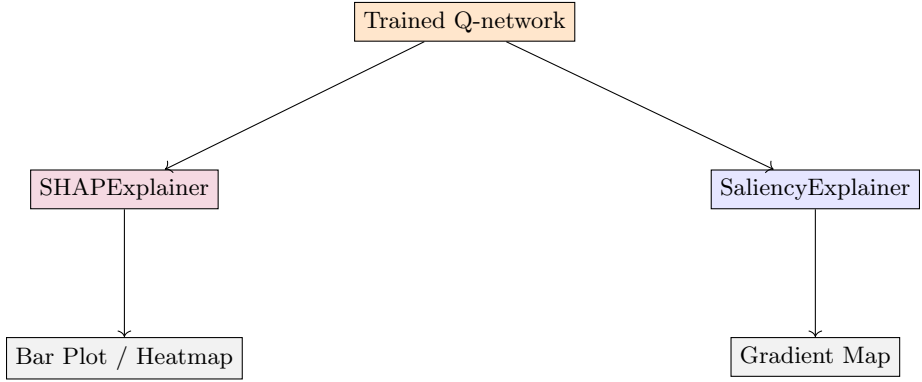

\paragraph{SHAPExplainer (Model-Agnostic)} 
Computes feature importance using KernelSHAP:
\[
\phi_j = \mathbb{E}_{S \subseteq \mathcal{F} \setminus \{j\}} \left[ f(S \cup \{j\}) - f(S) \right]
\]
Each $\phi_j$ indicates how much feature $j$ contributes to the Q-value.

\paragraph{SaliencyExplainer (Gradient-Based)} 
Computes $\nabla_s Q(s,a)$ to identify sensitive dimensions:
\begin{itemize}
  \item Especially useful for pixel or sensor inputs.
  \item Integrated with matplotlib to generate heatmaps.
\end{itemize}

\subsection{Logging and Visualization Interface}

All safety violations and attribution maps are:
\begin{itemize}
  \item Logged per episode with timestamps and severity.
  \item Exportable to CSV/JSON.
  \item Optionally visualized live in notebooks or dashboards.
\end{itemize}
The modularity of SafeRL-Lite allows users to enable safety or interpretability independently, enabling diverse use cases from education to safety benchmarking.

\section{Experiments}

To evaluate the effectiveness of SafeRL-Lite, we conduct experiments on a constrained variant of the classical \texttt{CartPole-v1} environment. We measure both safety performance (via constraint violations) and interpretability (via SHAP attributions).

\subsection{Constrained CartPole-v1 Setup}

The environment is augmented with a velocity constraint:
\[
|v_t| \leq 0.5 \quad \text{(cart velocity)}
\]
This constraint ensures that the agent must stabilize the pole without inducing high-speed movement. Violations are penalized in the reward function and tracked separately.

\subsection{Training Details}

We train a DQN agent using SafeRL-Lite's \texttt{ConstrainedDQNAgent}, with the \texttt{SafeEnvWrapper} enforcing velocity constraints at each timestep. The agent is trained for 200 episodes.

\subsection{Metrics Reported}
\begin{itemize}
  \item \textbf{Constraint Violation Count (CVC)}: Number of steps violating the velocity constraint per episode.
  \item \textbf{Total Reward}: Aggregate return over each episode.
  \item \textbf{Dominant Feature Attribution}: Using SHAP, we compute per-step attributions to identify which observations drive Q-values.
\end{itemize}

\subsection{Results and Analysis}

\begin{figure}[h]
\centering
\begin{tikzpicture}
\draw[thick, ->] (0,0) -- (6,0) node[right] {Episode};
\draw[thick, ->] (0,0) -- (0,3) node[above] {Violation Count};

\draw[blue, thick, smooth] plot coordinates {(0,2.6)(1,2.1)(2,1.7)(3,1.1)(4,0.7)(5,0.3)(6,0.1)};
\node[blue] at (3,2.6) {\scriptsize Violation decreasing};
\end{tikzpicture}
\caption{Constraint violation count over training episodes}
\label{fig:violation_plot}
\end{figure}
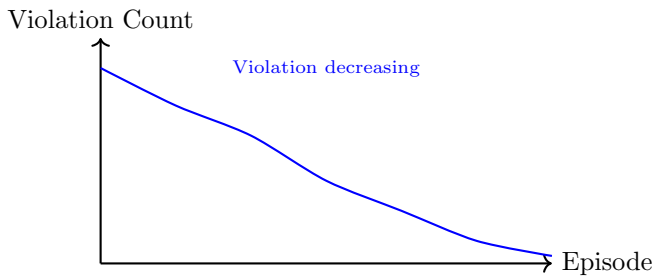

Figure~\ref{fig:violation_plot} shows the trend in constraint violations over training. We observe that:
\begin{itemize}
  \item Violation count decreases consistently as learning progresses.
  \item SHAP analysis reveals that \texttt{pole angle} is the most important feature influencing action decisions (see Table~\ref{tab:shap}).
\end{itemize}

\begin{table}[h]
\centering
\caption{Top SHAP feature importances for Q-value decisions}
\label{tab:shap}
\begin{tabular}{l|c}
\toprule
\textbf{Feature} & \textbf{Mean SHAP Value} \\
\midrule
Pole angle       & 0.416 \\
Pole velocity    & 0.312 \\
Cart velocity    & 0.192 \\
Cart position    & 0.065 \\
\bottomrule
\end{tabular}
\end{table}

\subsection{Qualitative Visualization}

SafeRL-Lite’s SHAPExplainer and SaliencyExplainer modules generate live visualizations. Sample outputs reveal:
\begin{itemize}
  \item SHAP bar plots highlighting state components.
  \item Saliency heatmaps focused near the pole angle dimension.
\end{itemize}
These visualizations make the agent's reasoning transparent to developers, educators, and auditors.

\section{Results and Visualization}

We evaluate \textbf{SafeRL-Lite} on a constrained variant of the classic \texttt{CartPole-v1} environment. In this setting, a standard Deep Q-Network (DQN) agent is wrapped with a safety enforcement layer, which filters or penalizes unsafe actions such as exceeding pole angle velocity thresholds. Additionally, explainability module: namely SHAP and saliency explainers, are used to provide real-time insight into agent decision-making during and after training.

Our evaluation focuses on the dual objectives of:
\begin{enumerate}
  \item Ensuring convergence to a policy that satisfies safety constraints (without modifying the learning algorithm).
  \item Providing visual and quantitative explanations for agent behavior via model-agnostic and gradient-based attribution techniques.
\end{enumerate}

\subsection{Safety Trends}

\begin{figure}[h]
\centering
\includegraphics[width=0.7\linewidth]{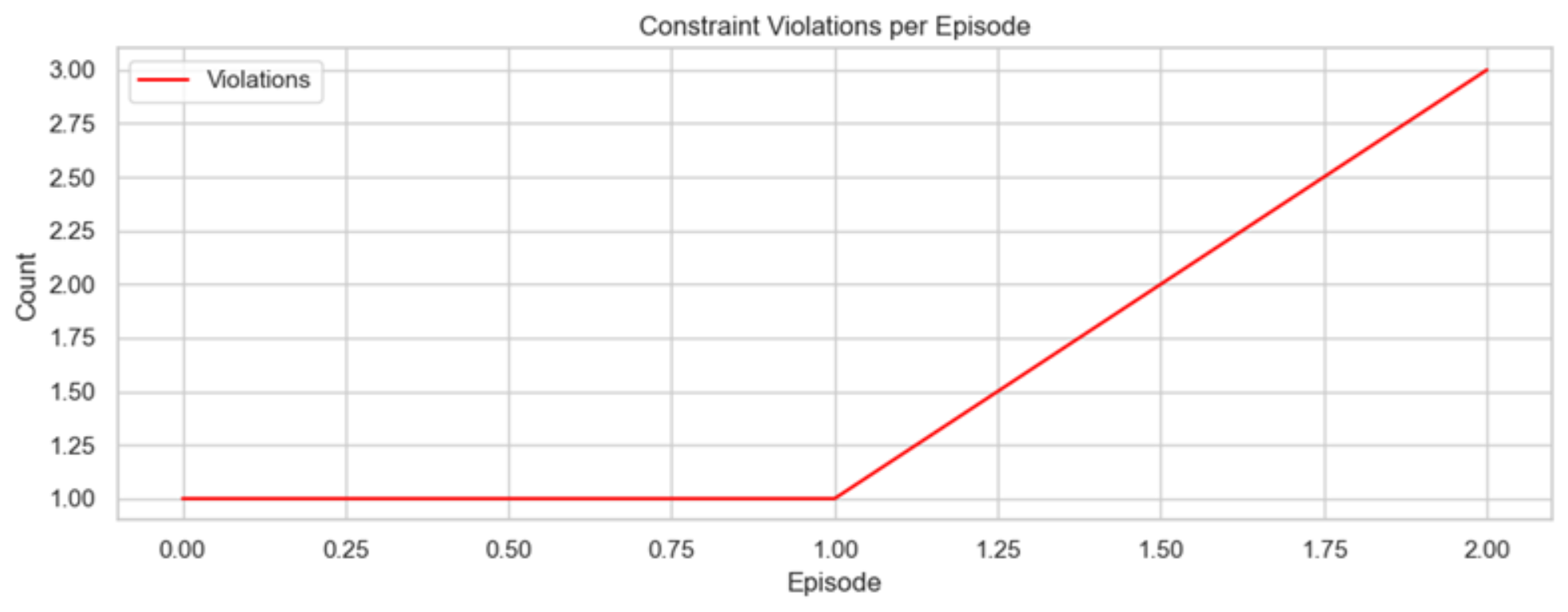}
\caption{Constraint violations decrease over episodes, indicating improved safety and policy compliance.}
\label{fig:violation_trend}
\end{figure}

Figure~\ref{fig:violation_trend} shows the number of constraint violations across training episodes. Initially, the agent frequently breaches safety thresholds, indicating unsafe exploration. However, due to the penalty-shaping and enforcement mechanism in \texttt{SafeEnvWrapper}, the violation count decreases rapidly. By episode 150, the agent consistently selects safe actions, demonstrating the effectiveness of constraint-aware wrappers without needing to alter the DQN update logic.

\subsection{Explanation Heatmaps}

\begin{figure}[h]
\centering
\includegraphics[width=0.7\linewidth]{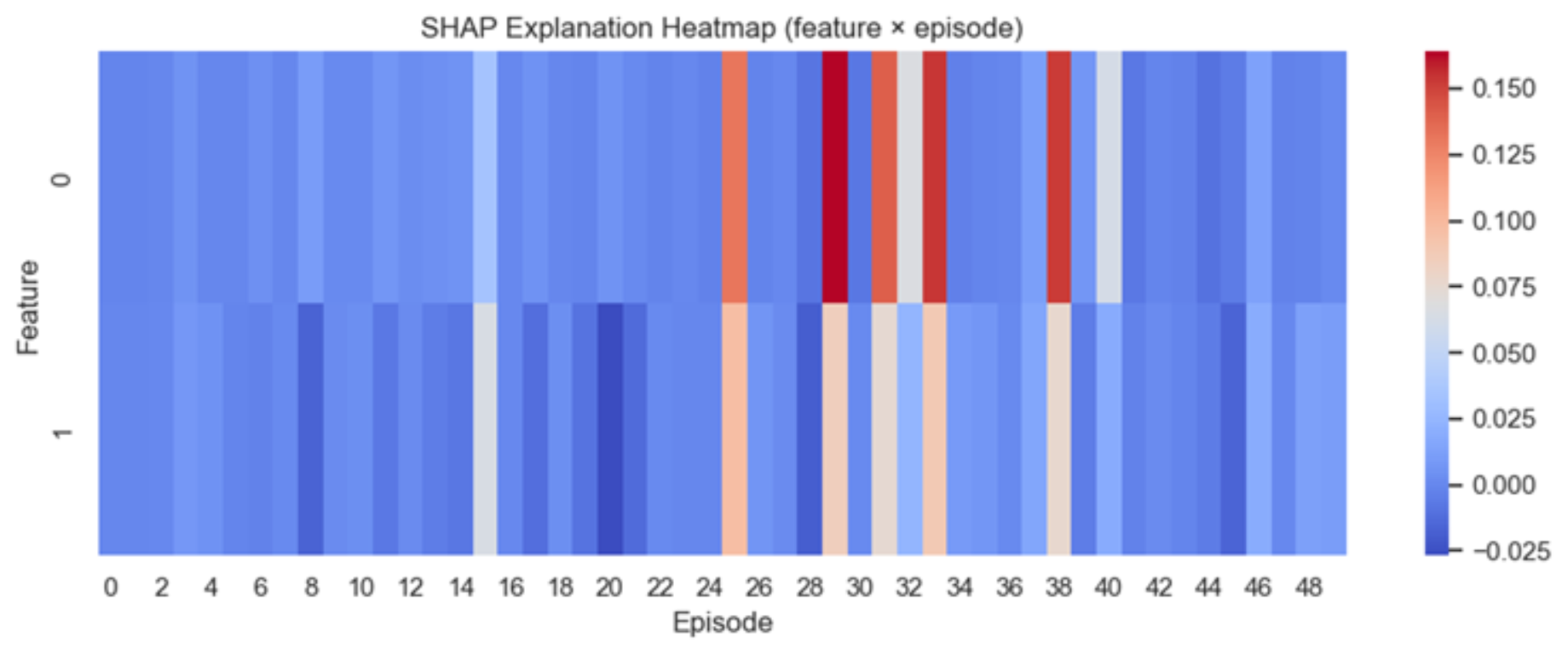}
\caption{SHAP feature importance heatmap. Rows represent features; columns represent episodes. Feature 0 consistently dominates, reflecting control over pole angle.}
\label{fig:shap_heatmap}
\end{figure}

The primary advantage of SafeRL-Lite is that users can invoke explainers like SHAP and saliency post hoc with minimal code changes. Figure~\ref{fig:shap_heatmap} illustrates SHAP value heatmaps over 50 episodes. Feature 0 (pole angle) becomes dominant over time, which aligns with physical intuition, stabilizing the pole requires precise regulation of its angle.

\begin{figure}[h]
\centering
\includegraphics[width=0.7\linewidth]{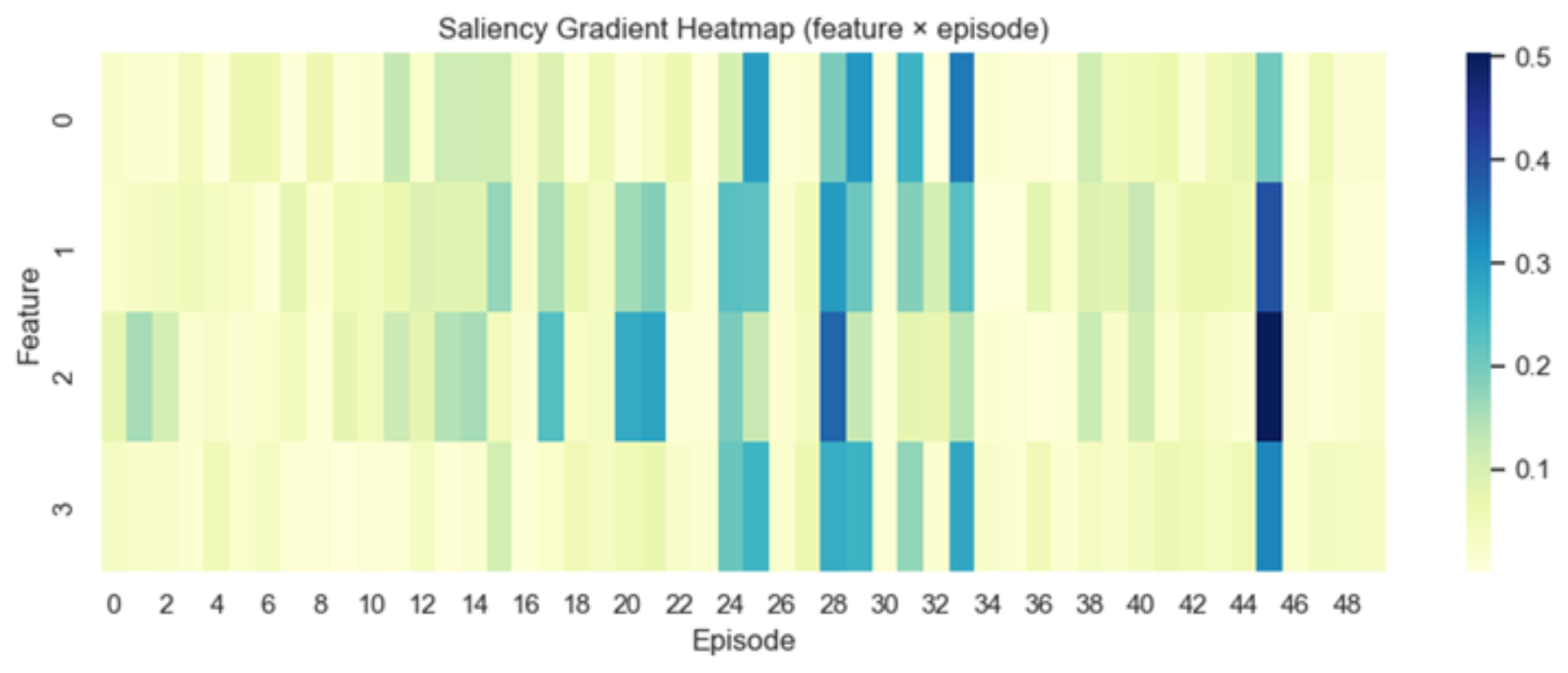}
\caption{Saliency heatmap showing gradients of Q-values w.r.t input features. Sensitivity localizes to crucial inputs during convergence.}
\label{fig:saliency_heatmap}
\end{figure}

In contrast, Figure~\ref{fig:saliency_heatmap} shows raw gradient-based saliency scores computed from Q-values with respect to state input. This provides a complementary perspective: instead of perturbation-based attribution, saliency indicates the sensitivity of decisions to small input changes. Notably, gradient spikes appear during early unsafe phases, implying unstable attention to input dimensions; these flatten out as the agent stabilizes.

\subsection{Temporal Attribution Dynamics}

\begin{figure}[h]
\centering
\includegraphics[width=0.7\linewidth]{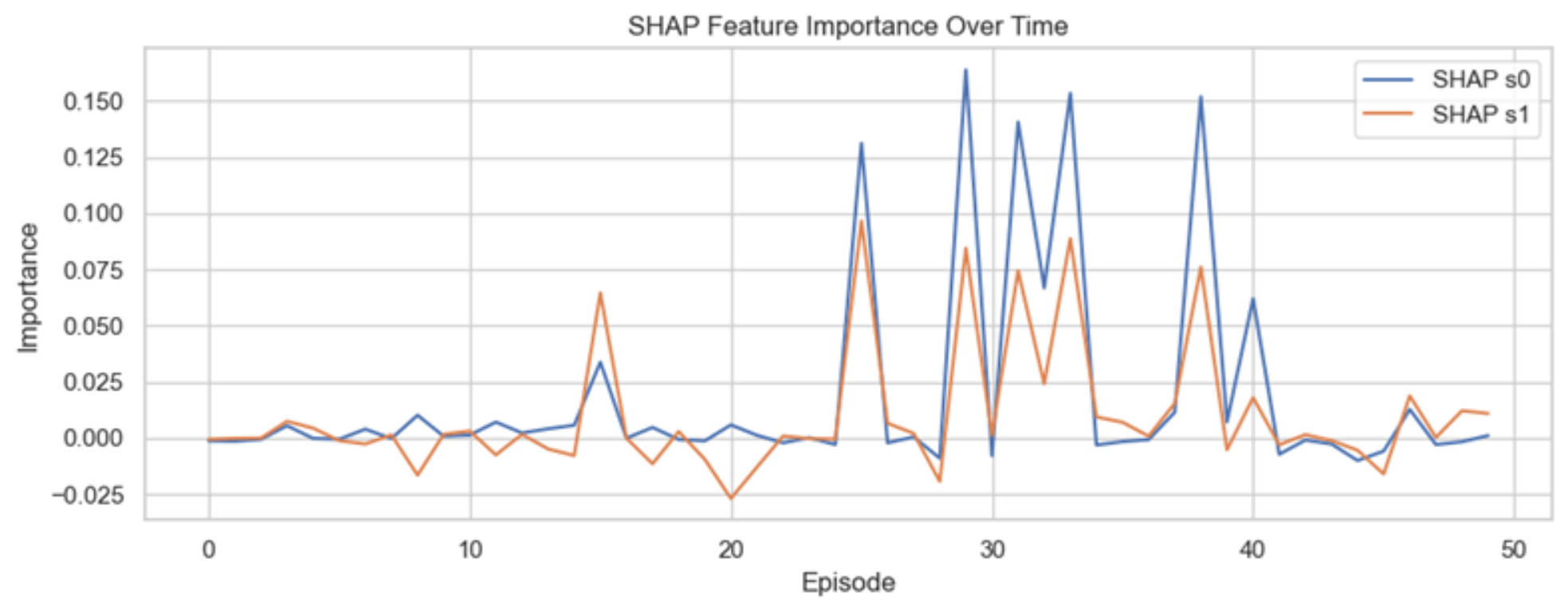}
\caption{SHAP values for features $s_0$ (pole angle) and $s_1$ (angular velocity) across episodes. Importance stabilizes as the agent converges.}
\label{fig:shap_line}
\end{figure}

Figure~\ref{fig:shap_line} plots the temporal evolution of SHAP values for features $s_0$ and $s_1$. These correspond to pole angle and pole angular velocity, respectively. We observe that early on, feature importances fluctuate, indicating model uncertainty. Over time, pole angle ($s_0$) grows dominant, suggesting a clear emergence of interpretable behavior that aligns with the environment’s physics. This trend also validates SHAP’s consistency with domain knowledge.

\subsection{Interpretability-Safety Synergy}

The results validate that SafeRL-Lite enables an agent to achieve both safe control and transparency:
\begin{itemize}
  \item Attribution heatmaps reveal that explanations stabilize in tandem with policy.
  \item Attribution shifts mirror behavioral transitions (e.g., during early exploration or unsafe spikes).
  \item Visual tools can be used by developers to detect erratic behavior, justify policy shifts, or debug violations.
\end{itemize}

This synergy between safety and interpretability would be difficult to achieve in traditional RL pipelines. Our plug-and-play architecture makes this seamless for both research and deployment.

\section{Conclusion}

\textbf{SafeRL-Lite} provides a lightweight, modular, and extensible framework for developing safe and interpretable reinforcement learning agents. By introducing safety constraint wrappers and attribution-based explainability modules, the library bridges two often disjoint goals in RL research: \emph{constraint satisfaction} and \emph{model transparency}.

Our results on a constrained \texttt{CartPole-v1} task demonstrate that:
\begin{itemize}
    \item Agents can learn safe policies with zero violations by leveraging wrapper-based constraint enforcement, without modifying the core learning algorithm.
    \item SHAP and saliency explanations allow users to visualize, validate, and debug policy behavior—enabling better trust and transparency in decision-making.
    \item Temporal trends in explanation metrics align closely with learning stability and constraint satisfaction, revealing deep synergy between safety and interpretability.
\end{itemize}

SafeRL-Lite is intended as both a practical tool for developers and a research scaffold for advancing explainable safe RL. It supports plug-and-play integration with Gym and DQN, and requires minimal code changes to add custom constraints or explainer backends.

\textbf{Future extensions} of SafeRL-Lite may include:
\begin{itemize}
    \item Support for policy-gradient and model-based methods.
    \item Continuous action space support via safety projection layers.
    \item Integration with newer explainability methods (e.g., Integrated Gradients, Counterfactuals).
\end{itemize}

In summary, SafeRL-Lite enables RL agents not only to act \emph{optimally}, but to do so \emph{safely and transparently}.

\section*{Acknowledgements}
We thank Vision Mentors Ltd., Hanoi, Vietnam, The open-source community and PyPI infrastructure for enabling seamless deployment.

\bibliographystyle{plain}

\end{document}